%% file: 00_main.tex
\documentclass[conference]{IEEEtran}
\IEEEoverridecommandlockouts
\usepackage{amsmath}
\usepackage{mathtools}
\usepackage{cite}
\usepackage[ruled, lined, linesnumbered, commentsnumbered, longend,noend]{algorithm2e}
\usepackage[font=normal,labelfont=bf]{caption}
\usepackage{hyperref}
\usepackage{amsmath,amssymb,amsfonts}
\usepackage{algorithmic}
\usepackage{graphicx}
\usepackage{textcomp}
\usepackage{xcolor}
\usepackage{breqn}

\def\BibTeX{{\rm B\kern-.05em{\sc i\kern-.025em b}\kern-.08em
    T\kern-.1667em\lower.7ex\hbox{E}\kern-.125emX}}
\definecolor{pink}{RGB}{255, 192, 203}

\newcommand{\systemName}{\textit{ExtPerFC}}
\newcommand{\citetk}[1]{{\cite{#1}}}
\newcommand{\argminI}{\mathop{\mathrm{argmin}}\nolimits} 
 
\usepackage{fancyhdr} 
\begin{document}

\title{\LARGE \bf \systemName: An Efficient 2D and 3D Perception\\ Software-Hardware Framework for Mobile Cobot}

\author{Tuan Dang, Khang Nguyen, and Manfred Huber\\[0.1cm]
University of Texas at Arlington\\[0.1cm]
\href{mailto:tuan.dang@uta.edu}{\texttt{tuan.dang@uta.edu}}, \href{mailto:khang.nguyen8@mavs.uta.edu}{\texttt{khang.nguyen8@mavs.uta.edu}}, \href{mailto:huber@cse.uta.edu}{\texttt{huber@cse.uta.edu}}}

\maketitle
\begin{abstract}
As the reliability of the robot's perception correlates with the number of integrated sensing modalities to tackle uncertainty, a practical solution to manage these sensors from different computers, operate them simultaneously, and maintain their real-time performance on the existing robotic system with minimal effort is needed. In this work, we present an end-to-end software-hardware framework, namely~\systemName, that supports both conventional hardware and software components and integrates machine learning object detectors without requiring an additional dedicated graphic processor unit (GPU). We first design our framework to achieve real-time performance on the existing robotic system, guarantee configuration optimization, and concentrate on code reusability. We then mathematically model and utilize our transfer learning strategies for 2D object detection and fuse them into depth images for 3D depth estimation. Lastly, we systematically test the proposed framework on the Baxter robot with two 7-DOF arms, a four-wheel mobility base, and an Intel RealSense D435i RGB-D camera. The results show that the robot achieves real-time performance while executing other tasks (\textit{e.g.}, map building, localization, navigation, object detection, arm moving, and grasping) simultaneously with available hardware like Intel onboard CPUS/GPUs on distributed computers. Also, to comprehensively control, program, and monitor the robot system, we design and introduce an end-user application. The source code is available at \url{https://github.com/tuantdang/perception_framework}.
\end{abstract}


\input{01_introduction}
\input{02_related_work}
\input{03_framework}
\input{04_map_nav_motion}
\input{05_perception}
\input{06_evaluation}
\input{07_limitations}
\input{08_conclusions}
\input{09_acknowledgement}

\bibliographystyle{IEEEtran}


\end{document}

%% file: 01_introduction.tex
\section{Introduction}
The past few years have seen an increasing number of different sensing modalities integrated into robots that significantly enhance robot perception, especially for autonomous service mobile cobots to perform map building, localization, navigation, object detection, arm moving, and efficiently grasping objects for safety purposes \citetk{hsiao2009reactive}. However, besides support for conventional tasks in robot control and navigation, efficient techniques to deal with 2D and 3D perception requiring expensive computational power must also be deployed on the same system. For this reason, an efficient software-hardware framework that concurrently enables sensors, communication, perception, navigation, and motion planning to operate seamlessly is necessary for incremental robotic development.

Previous works focus on only one of the aspects mentioned above \citetk{hmedan2022adapting, vice2022leveraging}, and flexible component integration is often omitted. Recently, Robot Operating System 2 (ROS 2) \citetk{macenski2022robot} has improved security and reliability, which are critical criteria in a commercial product; meanwhile, ROS 1 is still popular among research communities and industry. Nevertheless, the single failure point of the ROS master causes poor performance if multiple sensing modalities are initiated simultaneously, especially with high bandwidth data in LiDAR sensors and RGB-D cameras. Moreover, integrating state-of-the-art machine learning (ML) with optimal configurations into the current ROS software stack can burden developers as no official framework can handle this task.

\begin{figure}[t!]
    \centering
    \includegraphics[width=1.00\linewidth]{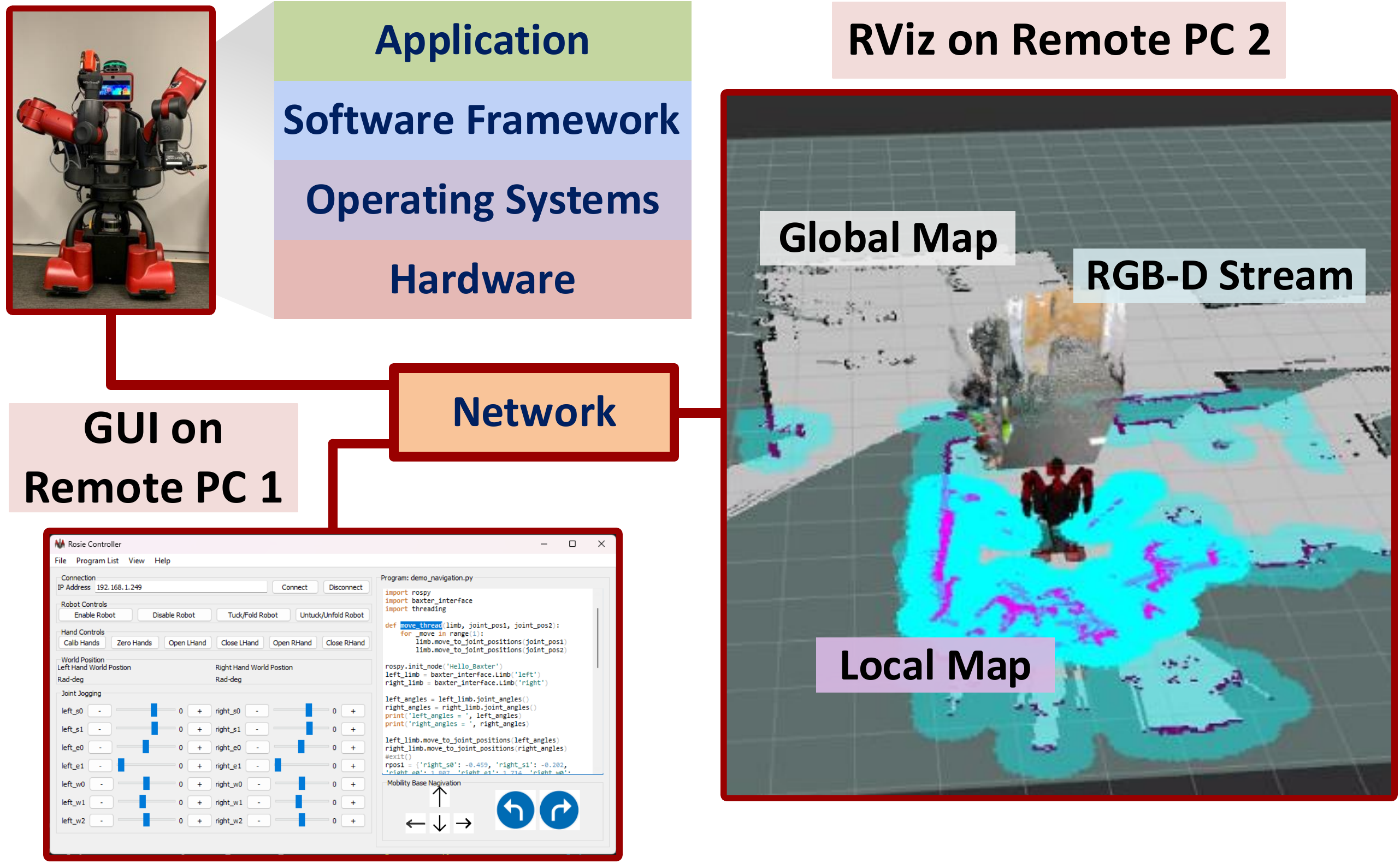}
    \caption{\systemName's concept for mobile cobot perception.}
    \label{fig:concept}
    \vspace{-15pt}
\end{figure} 

Recent works on deploying Deep Neural Networks (DNN) for a safe and secure automation domain \citetk{biondi2019safe,nazarova2021cobotar} propose a visionary hypervisor-centric architecture. Yet, the integration of ML modules is meticulously tailored for specific applications. Also, their complexity worsens when they are deployed on different computing hardware. Thus, a framework is needed to eliminate repetitive tasks (training, testing, and detection) and be compatible with available hardware on robot systems.

\begin{figure*}[t]
    \includegraphics[width=1.00\textwidth]{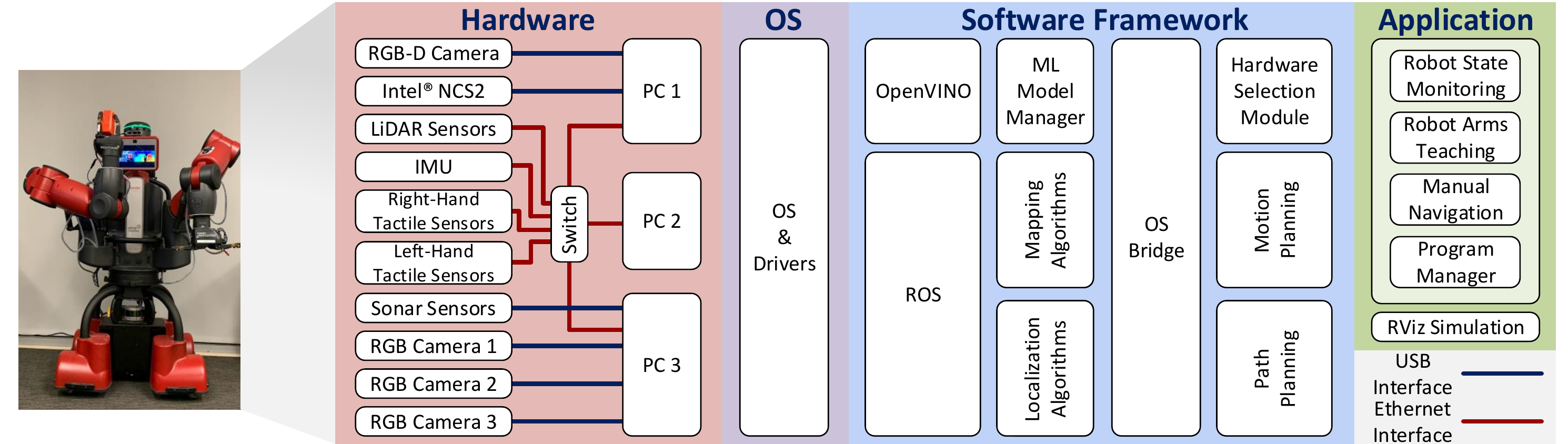}
    \centering
    \caption{\systemName~includes a hardware layer, an OS layer, a software framework, and an application layer.}
    \vspace{-15pt}
    \label{fig:framework_overview}
\end{figure*}

To fill this gap, we build an efficient hardware-software framework (Fig. \ref{fig:concept}) that allows simple integration of various tasks and different hardware driver versions. More importantly, we introduce a design that can support multiple state-of-the-art object detection models and execute them on low-end commodity devices in real time, enabling developers to manage computing tasks on available hardware in the system with high flexibility and minimum effort  (i.e., choose a set of devices to process data from sensors and deploy ML models).

In this work, the contributions are (1) building a complete software-hardware framework for a mobile cobot system that supports map building, localization, navigation, and motion planning (Sec. \ref{sec:map_nav_motion}), as well as 2D and 3D perception using state-of-the-art DNNs (Sec. \ref{sec:perception}), (2) verifying the framework feasibility and performance on a real robot system, and (3) producing a fast method to train multiple object detection using transfer learning (Sec. \ref{sec:transfer_od}) with open-source code.

However, to build such an efficient framework for 2D and 3D perception and other robot tasks, we have to overcome  challenges: (1) integrating a large amount of software library while guaranteeing dependencies, (2) building and evaluating multiple DNN models, and (3) solving incompatible hardware issues due to high variance in hardware selection. This work extends the previous preliminary work \cite{dang2023perfc} by overcoming the issue of finding the best configuration between available sensors. Furthermore, this work also focuses on building and testing 2D/3D perception more extensively. 

%% file: 02_related_work.tex
\section{Related Work}

\indent \textbf{Architecture of Robot Software:} Many software architectures are proposed and designed for industrial robot applications \citetk{rendiniello2020flexible} to cope with issues of robot-language dependency. Although they solve the issue, they incidentally limit users from developing new functions. Specifically, adding more sensing modalities or additional hardware becomes complicated since developers only access services from specific robot software development toolkits. Accessing services into the OS layer or other libraries is restricted. Moreover, using robot-language dependency prevents developers from accessing many open-source libraries. For those reasons, we develop a software framework that enables developers to flexibly access multiple system layers and access open-source libraries while maintaining simple integration with most state-of-the-art ML models. 

\indent{\textbf{Robot Mapping, Localization, and Navigation:}} Previous works on the Baxter robot \citetk{qureshi2019motion, pinto2016supersizing} primarily concentrate on picking and placing and motion planning tasks. Thus, the integration of LiDAR sensors along with the mobility base into the Baxter is crucial for its map building and navigation. Moreover, autonomous mobile service robots in the past with both navigation and tracking modules \citetk{veloso2015cobots, bellotto2008multisensor} are embedded with lightweight localization and mapping algorithms due to their application simplicity (i.e., specific tasks and object tracking in a known environment), which makes these robots difficult to expand and adjust to developers. In this work, we introduce an expandable and user-adjustable software-hardware framework for mobile cobots with state-of-the-art laser-based SLAM algorithms deployed. 

\indent{\textbf{Motion Planning:}} The main objective of arm motion planning is to find a trajectory from the end-effector position to desired positions, avoiding collision and minimizing the path cost and time complexity under the constraint that every point in the trajectory must have an inverse kinematic solution. In dealing with efficient arm motion planning, many methods have been used \citetk{kingston2019exploring, ichnowski2020gomp}, such as AtlasRRT and CBIRRT2. Most of them are well-supported by MoveIt! \citetk{chitta2012moveit}. Therefore, we reuse them for our framework implementation.

\indent{\textbf{Robot Perception:}} 2D perception \citetk{liu2016ssd, wang2023yolov7} is widely used in research and industrial applications, while 3D perception \citetk{mao20223d, qi2017pointnet} is dominating in autonomous driving vehicles with LiDAR sensors support, but limited in everyday object perception regarding the robotic domain. The fact of lack of robotic research in 3D perception is the absence of diverse labeled datasets since most of them are not specifically for robotic applications. In this work, we used a hybrid method of 2D state-of-the-art detection and 3D estimation methods. 

%% file: 03_framework.tex
\section{Software-Hardware Framework}

\subsection{Design Goals}
In order to build a more reliable software-hardware framework for mobile cobots, we define our design goals are: 
\begin{itemize}
    \item Reusability and simple integration into systems with mixed versions of OS and middleware: Linux \& ROS.
    \item Distributed computing with load balancing awareness.
    \item State-of-the-art ML models deployed with optimal configurations of cameras and processing devices.
\end{itemize}

Hardware is often compatible with a specific Linux kernel, and a specific ROS   distribution is only provided to a specific Linux version. For this reason, selecting hardware concurrently with selecting Linux kernels implies narrowing down options in choosing a ROS distribution for developers. Unfortunately, not all hardware works well with the same Linux kernel, leading to using various ROS versions in the same system. Therefore, calling the same APIs from different ROS distributions may cause backward incompatibility issues due to a slight change in the function prototype and the underlying implementation of that supported API.  We adopt message conversion \citetk{kim_gateway} between multiple communication protocols to implement the message translator between ROS versions.

The driver incompatibility problems can be solved by using suitable Linux kernels supporting these devices' drivers. However, it may raise backward incompatibility between a certain ROS distribution and APIs from other ROS distributions. Indeed, we encountered these compatibility issues with the Baxter robot (Ubuntu 14.04 and ROS Indigo) when executing motion APIs on Linux machines with Ubuntu versions other than Ubuntu 14.04. Therefore, an OS bridge between APIs from different ROS distributions is needed.

With the increasing requirements of computational tasks, a single computer system may no longer fit into a robotic application since the failure of that computer would cause the entire system to crash. We adopt distributed computing with load balance awareness between computers by periodically broadcasting the status to their peers. With this scheme, we can obtain load balancing at run-time. Two aspects must be considered for robots with multiple vision cameras: (1) bandwidth to acquire data and (2) processing power to perform detection or recognition tasks. We propose a method for selecting the optimal configuration at run-time for the combination of cameras, ML models, and devices.

\begin{algorithm}[t]
    \caption{\small Hardware Selection Algorithm}
    \label{alg:hardware_selection}
    
    \begin{small}
        \DontPrintSemicolon
        \SetKwInOut{KwAss}{Assume}
        \SetKwInOut{KwIn}{Input}
        \SetKwInOut{KwOut}{Output}
        \SetKwFunction{FMain}{hardware_selection}
        \SetKwProg{Pn}{function}{}{}
        
        \KwIn{S $= \{s_1, s_2, ..., s_n\} :=$ available sensors\\ 
            D $= \{d_1, d_2, ..., d_m\} :=$ available devices\\
        }
        
        \KwOut{
            C $= \{c_1, c_2, ..., c_n\} :=$ optimal configurations
        }

        
        \Pn{\FMain{S, D}}{ 
            \If{$|S|$ = 0 \text{\textbf{or}} $|D|$ = 0}{ 
                \textbf{return} C = \{$c_1$, $c_2$, ..., $c_n$\}
            }
            S = \texttt{sort\_by\_image\_size}(S)\\
            D = \texttt{sort\_by\_comp\_power}(D)\\
            \For{$s_i \in$ S}{
                \For{$d_j \in$ D}{
                    \If{(enet($s_i$, $d_j$) \text{\textbf{or}} usb($s_i$, $d_j$)) = 1}{ 
                        $c_i$ = ($s_i$, $d_j$)\\
                        \texttt{delete\_from\_list}($s_i$, S)\\
                        \texttt{delete\_from\_list}($d_j$, D)
                    }
                    \textbf{el}\If{enet($s_i$, $d_j$) = 0 \text{\textbf{and}} usb($s_i$, $d_j$) = 0}{
                        $d_k$ = \texttt{find\_connected}($s_i$, D$\backslash$\{$d_j$\})\\
                        $c_i$ = ($s_i$, $d_j$, $d_k$)\\
                        \texttt{delete\_from\_list}($s_i$, S)\\
                        \texttt{delete\_from\_list}([$d_j$, $d_k$], D)
                    }
                }
            }
            \texttt{hardware_selection}(S, D)
        }
        \KwRet C = \{$c_1$, $c_2$, ..., $c_n$\}
    \end{small}
\end{algorithm}

\subsection{Framework Description}
\begin{figure*}[t]
    \includegraphics[width=1.00\textwidth]{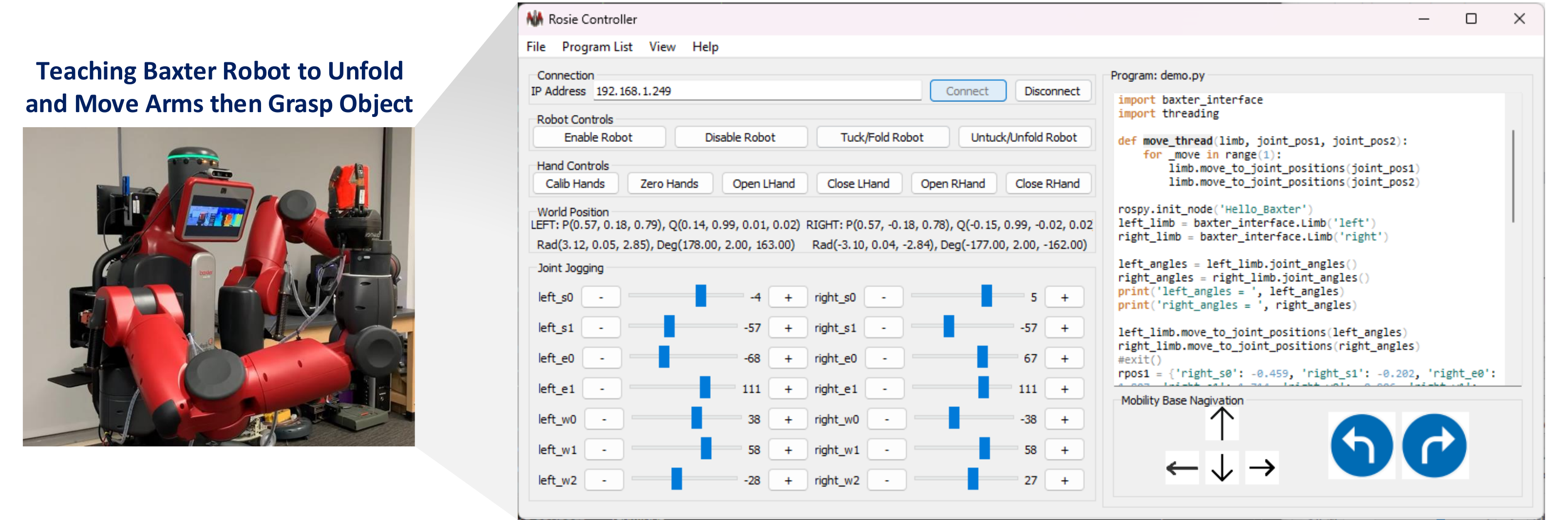}
    \centering
    \caption{Graphic User Interface (GUI) and Program Manager provide robot teaching, monitoring, navigation, and programming.}
    \vspace{-15pt}
    \label{fig:gui}
\end{figure*}

As illustrated in Fig. \ref{fig:framework_overview}, we structure the software-hardware framework as four different layers: hardware layer, OS layer, software framework, and application layer.
\begin{enumerate}
    \item \textbf{Hardware Layer} includes distributed computers, sensors (cameras, tactile sensors, LiDAR sensors), IMU, actuators, and ML accelerators and how they interface with each other by USB and Ethernet ports.
    \item \textbf{OS Layer} contains an OS and device drivers that support connecting devices at the hardware layer. 
    \item \textbf{Software Framework} is the core contribution of this work. It bridges different Linux and ROS distributions, selects optimal configurations of sensors and ML models to available hardware to operate in real-time, and supports basic functionalities of the robot system, including map building, localization, navigation, motion planning, and arm moving for grasping. Most of the components are implemented on top of ROS and OpenVINO.
    \item \textbf{Application Layer} allows users to control and monitor the robot via a GUI, as shown in Fig. \ref{fig:gui}, which is written using the PyQt5 toolkit. The users can manually control the robot's arms and joint positions, navigate the mobility base, and program the robot using Python. Also, native simulations such as RViz can be used to monitor the robot.
\end{enumerate}

\begin{figure}[t]
    \centering
    \includegraphics[width=1.00\linewidth]{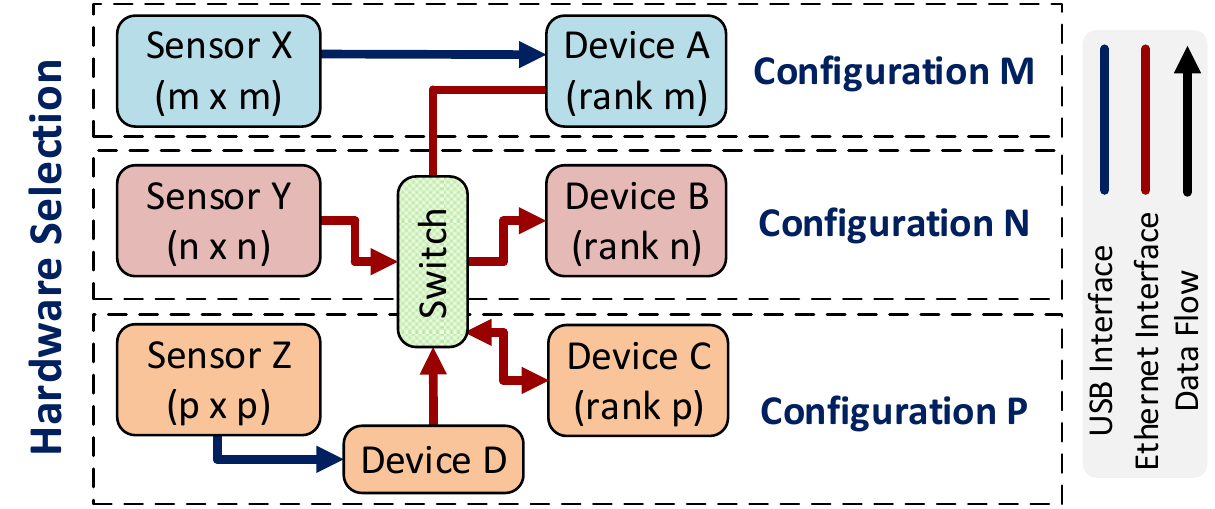}
    \caption{Illustration of the hardware selection algorithm.}
    \vspace{-15pt}
    \label{fig:hardware_selection_algo}
\end{figure}

\subsection{Hardware Configuration Optimization}
\textbf{Problem Formalization}: We need to distribute processing of data from $n$ sensors $S = \{s_1, s_2, ..., s_n\}$ into the distributed computers. Each distributed computer could contain a CPU, a GPU, and a VPU, and we denote these processing units as $ d_i$s in $D$. The image stream acquired from a sensor comes from either a USB bus if that sensor is connected to a device via a USB port or from the network via an Ethernet switch, in which a device may ask the neighbor device to share the image stream. Also, we define each configuration as $c_i = [s_i, d_j]$, where $s_i$ is from $S$ and $d_j$ is from $D$.

\indent \textbf{Explanation \& Description}: To allocate the most suitable resources (devices) to the inputs (sensors) in a best-fit manner as in \citetk{dang2022iotree}, we prioritize the following criterion: (1) a higher-ranked device in terms of computational power should process a sensor that acquires higher input image sizes, and make the following constraints in correspondence to the hardware layer Fig. \ref{fig:framework_overview}, (2) any sensor is interfaced to at least one device by only either Ethernet or USB ports, and (3) any pair of devices are able to share data with each other via Ethernet ports. 

We first sort both $S$ and $D$ lists in descending order in terms of input image sizes and computational power, respectively, as depicted in \texttt{sort_by_image_size(S)} and \texttt{sort_by_comp_power(D)}. We then establish configurations for sensors and devices by nested-looping through $S$ and $D$ and check the following booleans, as described below:
\begin{equation*}
    \texttt{usb(s$_i$,d$_j$)} =
    \begin{cases}
      1, \text{ if $s_i$ and $d_j$ are connected via USB}\\
      0, \text{ otherwise}
    \end{cases}    
\end{equation*} 
\begin{equation*}
    \texttt{enet(s$_i$,d$_j$)} =
    \begin{cases}
      1, \text{ if $s_i$ and $d_j$ are connected via ENET}\\
      0, \text{ otherwise}
    \end{cases}    
\end{equation*} 

If sensor $i$ interfaces to device $j$ via Ethernet or USB, we can group them as one configuration, as configuration $M$ (with USB port interface) and configuration $N$ (with Ethernet interface) in Fig. \ref{fig:hardware_selection_algo}. However, if the sensor $i$ does not directly connect to the corresponding device $j$, we will find an already-connected device as an intermediate device $k$ to share the sensing data stream with device $j$. Thus, in this case, we group that sensor with the processing device $j$ and the intermediate device $k$ as one configuration as configuration $P$ in Fig. \ref{fig:hardware_selection_algo}. 

Note that after completing grouping each configuration, we also have to remove the configured elements from $S$ and $D$ lists: \texttt{delete_from_list(s$_i$, S)} and \texttt{delete_from_list(d$_j$, D)}, and hence recursively call the \texttt{hardware_selection(S, D)} function until either $S$ or $D$ list is empty. The pseudocode of the algorithm is presented in Alg. \ref{alg:hardware_selection}.

%% file: 04_map_nav_motion.tex
\section{Robot Mapping, Localization, Navigation, and Arm Motion Planning}
\label{sec:map_nav_motion}

\subsection{Robot Mapping and Localization}
Deploying the robot in an unknown dynamic environment with high uncertainty in robot perception (e.g., cameras, LiDAR sensors) requires a robust simultaneous mapping and localization (SLAM) algorithm to attain the highest map-building and localization accuracy. Moreover, building an environment map and localizing within that map in a robot system is essential to enabling path planning and path execution and later avoiding obstacles in robot navigation. In the Baxter robot, we examine the reconstructed maps by using various ROS-based SLAM algorithms on the VLP-16 LiDAR sensors and prioritize GMapping for its highest accuracy and availability in multiple ROS versions.

The main goal of GMapping is to compute the posterior over maps and trajectories $p(x_{1:t}, m\text{ }|\text{ }z_{1:t}, u_{0:t})$ given the estimated posterior $p(x_{1:t}\text{ }|\text{ }z_{1:t}, u_{0:t})$, where $x_{1:t}$ are robot poses up to time $t$, $z_{1:t}$ is the observation up to time $t$, and $u_{0:t}$ is the control sequence. The GMapping is modeled as:
\begin{equation*}
    p(x_{1:t}, m\text{ }|\text{ }z_{1:t}, u_{0:t}) = p(m\text{ }|\text{ }z_{1:t}, x_{1:t}) \cdot p(x_{1:t}\text{ }|\text{ }z_{1:t}, u_{0:t})
\end{equation*}

In specific, the posterior  $p(x_{1:t}\text{ }|\text{ }z_{1:t}, u_{0:t})$ is treated as a particle that is associated with a proposal map using the Rao-Blackwellized particle filter (RBPF). Meanwhile, the posterior over maps $p(m\text{ }|\text{ }z_{1:t}, x_{1:t})$ is built given the robot poses $x_{1:t}$ and observations $z_{1:t}$. As the robot obtains observations and control measurements, the RBPF for GMapping is iterative, repeating the following steps: (i) sampling, (ii) importance weighting, (iii) resampling, and (iv) map estimating \citetk{grisetti2005improving}.

\subsection{Robot Navigation}
To navigate the robot in an environment, we reconstruct the global cost map and local cost map, where obstacles appear. The trajectory from source to destination is generated while minimizing the path cost in both maps. In this work, we inherit these ROS-based features and translate them into our control command in the application layer (our APIs).

\subsection{Arm Motion Planning}
The main objective of arm motion planning is to find a trajectory from the end-effector source position to the desired positions avoiding collision and minimizing the path cost and time complexity under the constraint that every point in the trajectory must have an inverse kinematic solution. In dealing with efficient arm motion planning, many methods have been used \citetk{kingston2019exploring, ichnowski2020gomp}, such as AtlasRRT and CBIRRT2. Most of them are well-supported by MoveIt!. Thus, we reuse them for the implementation of our work.

\subsection{Graphic User Interface \& Program Manager}
To control the robot, we not only command from one computer but also command from other computers for synchronization tasks. Furthermore, working on multiple computers requires considerable steps like login using SSH, editing source code, and uploading and executing code. These steps are repetitive and time-consuming, becoming a burden for developers. We build a graphic user interface (GUI) application, as shown in Fig. \ref{fig:gui}, in this framework to eliminate these burdens while developing robotic software. 

The key idea is that this GUI application can run on a remote computer (Windows or Linux) and is able to connect to available on the robot so that it can execute specific tasks on each computer remotely. For example, tucking and untucking robots must run programs on the Bater robot, while moving the arm from one position to another can be conducted on another computer. Yet, another advanced tool is the program manager, allowing developers to edit Python code on a remote computer and execute their own customized code. This GUI application eventually saves the programming time of developers by getting rid of repetitive routines, thus making robotic software development more productive.

To allow the remote GUI applications to execute a command or a Python script, a gateway on the robot system is needed. This gateway talks directly to other computers via ROS message and ROS bridge adaption layer. In specific, it operates as a translator between the GUI application and the entire robot system by translating the commands from GUI to robot system commands and delivering commands to specific computers. In the meantime, it acquires all information about the robot system from distributed computers and periodically sends it to the connected remote computer.

%% file: 05_perception.tex
\section{Robot Perception}
\label{sec:perception}

\subsection{Transfer Learning for 2D Object Detectors}
\label{sec:transfer_od}

To perform real-time detection and recognition of objects with a limited richness compared with those in open datasets like ImageNet and MS COCO datasets, we adopt two strategies: transfer learning \citetk{zhuang2020comprehensive} and single state detection \citetk{liu2016ssd}. We first transfer knowledge from a rich domain feature into a sparse domain feature, which represents our dataset. Herein, we then mathematically model the overall concept of transfer learning and address the two questions: (i) what to be transferred between models and (ii) how to transfer that knowledge.

\begin{figure}[t]
    \centering
    \includegraphics[width=1.00\linewidth]{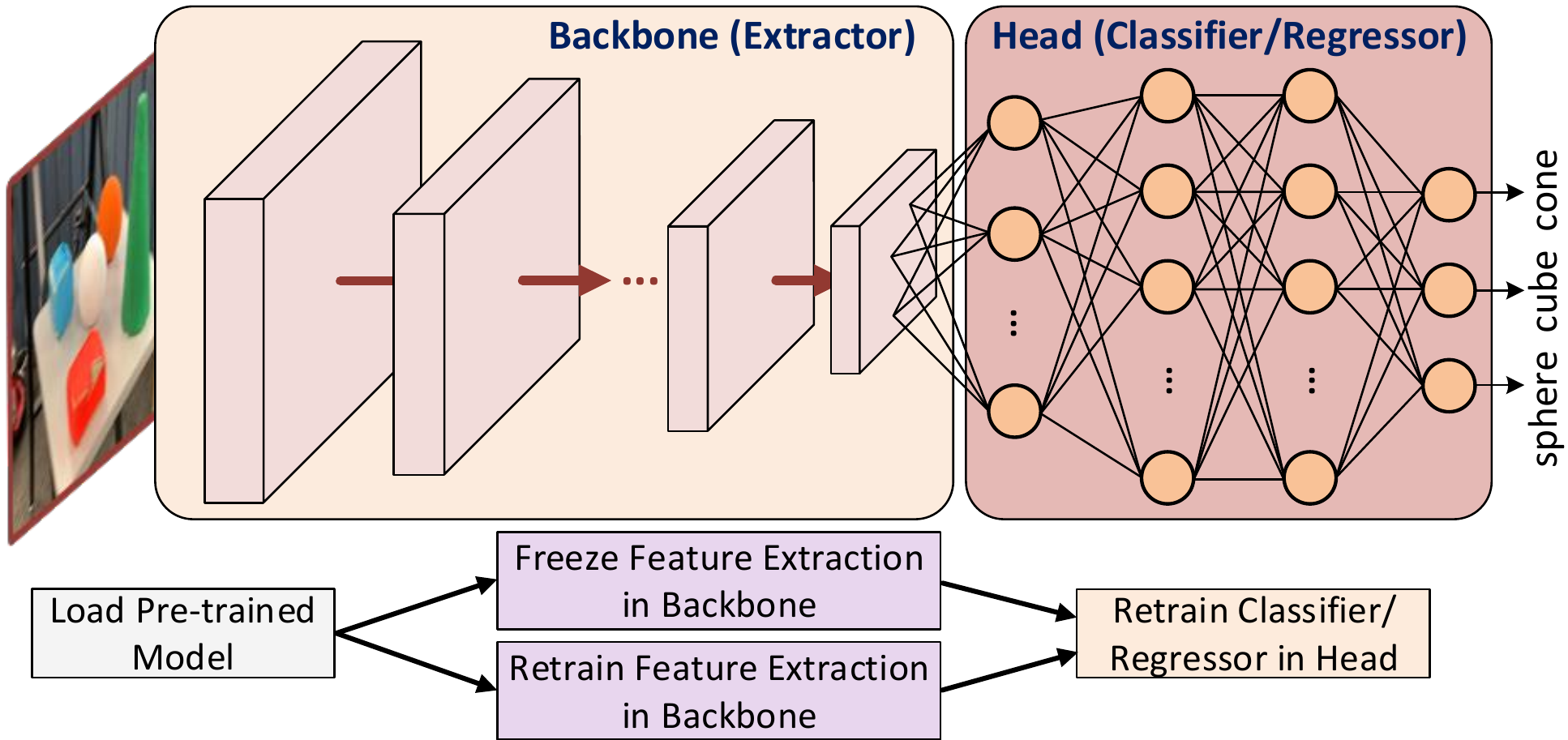}
     \caption{Illustration of transfer learning strategies: instance (fine-tuning) and feature representation transfer learning.}
    \vspace{-15pt}
    \label{fig:transfer_learning_strategies}
\end{figure}

\indent \textbf{Transfer Learning Formulation}: A domain is defined by $\mathcal{D}=\{ \mathcal{X}, P(X) \}$ where $X = \{x_1, x_2,..\} \in \mathcal{X} $ with $\mathcal{X}$ representing the feature space, and $P(X)$ its marginal distribution. Let $\mathcal{T} = \{\mathcal{Y}, P(Y|X)\}$ be the learning task that learns from training pairs $(x_i,y_i)$ with $y_i \in \mathcal{Y}$ in the label space. The objective of transfer learning is to improve the predictive function $P(Y_t|X_t)$ in target domain $\mathcal{D}_t=\{\mathcal{X}_t, P(X_t)\}$ using knowledge in the source domain $\mathcal{D}_s=\{ \mathcal{X}_s, P(X_s) \}$ and source learning task $\mathcal{T}_s = \{\mathcal{Y}_s, P(Y_s|X_s)\}$. 

Let $P(Y|X)=f(X, \beta)$ where $f$ is the task function. The minimizer for the trainable parameters, $\beta$, is written in terms of the loss function, $L(\cdot, \cdot)$, and the task function, $f$, as follows: 
\begin{equation}
    \argminI_{\beta}\sum_{X}L\left[f\left(X,\beta\right), Y\right]
\end{equation}

With respect to DL and computer vision concepts, we divide the task function into two components: feature extraction (backbone) and detection (head), such that $f(X,\beta)=(f^D \circ f^F)(X, \beta^D, \beta^F)$ where $f^D$ and $f^F$ are task function for detection and feature extraction, respectively, and $\beta^D$, $\beta^F$ are parameters for detection and feature extraction, respectively. The analogous minimizer for $\beta_t^F$ and $\beta_t^D$ is:
\begin{equation}
    \argminI_{\left\{\beta_t^D, \beta_t^F\right\}} \sum_{X_t}L\left[\left(f_t^D \circ f_t^F\right) \left(X_t, \beta_t^D, \beta_t^F\right), Y_t\right]
    \label{eq:transfer_learning}
\end{equation}

Since features in the source domain are more generalized and sufficiently cover our target domain, we assume that the feature space in the source domain and target domain are similar. However, our target labels are different ($\mathcal{Y}_s \neq \mathcal{Y}_t$) since we retrain the models with in-lab objects (cone, cube, and sphere). Here, we utilize two transfer learning strategies: (1) instance transfer, where the marginal distribution of source features is different from that of target features, and (2) feature representation transfer, where we fit the source feature domain into the target feature domain (Fig. \ref{fig:transfer_learning_strategies}). 

To implement instance transfer, we transfer $\left(\beta_s^D, \beta_s^F\right) \rightarrow \left(\beta_t^D, \beta_t^F\right)$, where $\beta_s^D$ and $\beta_s^F$ are resultants from source task functions, and fine-tune $\left(\beta_t^D, \beta_t^F\right)$ using Eq. \ref{eq:transfer_learning}. For feature representation transfer, we separate the source task into two components (backbone and head) and transfer the entire source task's backbone into the target task. In specific, we transfer $\beta_s^F \rightarrow \beta_t^F$, and train $\beta_t^D$ using Eq. \ref{eq:transfer_learning}. We also train with randomly initialized weights as a third strategy for accuracy comparisons in Sec. \ref{sec:eval}.

\begin{figure}[t]
    \centering
    \includegraphics[width=1.00\linewidth]{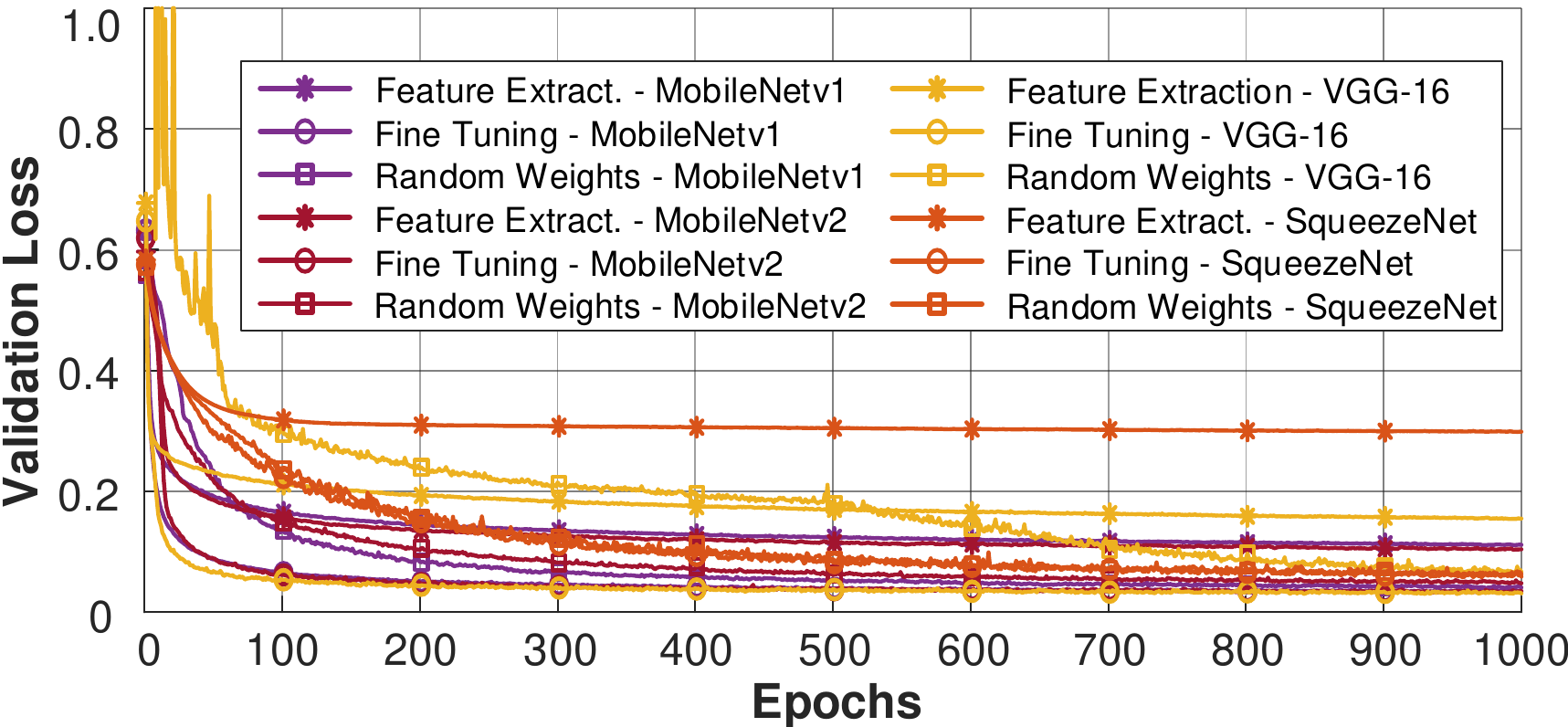}
    \caption{Validation losses with different transfer learning strategies and random weights training on MobileNetv1, MobileNetv2 Lite, VGG-16, and SqueezeNet models.}
    \label{fig:valid_loss}
    \vspace{-15pt}
\end{figure}

\subsection{Depth Estimation}
\label{sec:3d_depth_est}

We obtain depth images and RGB images simultaneously from the Intel RealSense D435i camera, which also well-handles the depth image creation process, including camera calibration, image rectification, and disparity computation. 

\begin{figure*}[t]
    \includegraphics[width=1.00\textwidth]{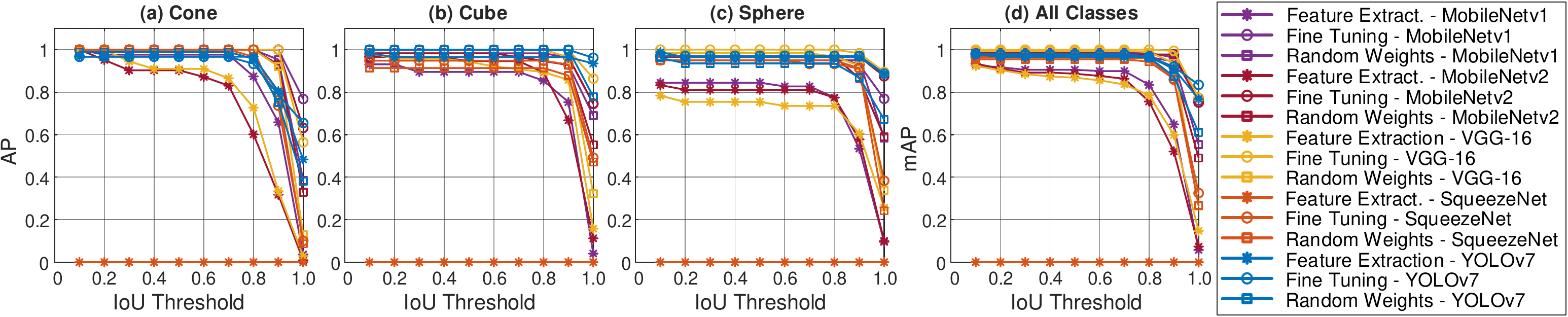}
    \centering
    \caption{AP for each class and mAP among all classes on MobileNetv1, MobileNetv2 Lite, VGG-16, SqueezeNet, and YOLOv7-tiny models. All models are trained with three different strategies, as described in Sec. \ref{sec:perception}.}
    \vspace{-15pt}
    \label{fig:ap_and_map}
\end{figure*}

As the whole predicted bounding box also covers non-detected objects, we average the depth of the bounding box would incur estimation errors. We, therefore, scale the bounding box small enough at the center of the box, then calculate the estimated depth of the object, $D$, by averaging depth values of each pixel in the scaled region as follows: 

\begin{equation}
    D = (w \times h)^{-1} \left[\sum_{i = x_0 - w/2}^{x_0 + w/2}\sum_{j = y_0 - h/2}^{y_0 + h/2}d(i, j) \right]
    \label{eq:depth_estimation}
\end{equation}

\noindent where $d(i, j)$ returns the depth value at pixel $(i, j)$, $w$ and $h$ indicate the width and height of the scaled region, respectively, and $(x_0,y_0)$ are center coordinates of the bounding box. 

Note that, in this framework, we choose the scaled region as $20 \times 20$ pixels to guarantee optimal selection for computational power while maintaining the correctness of depth estimation.

%% file: 06_evaluation.tex
\section{Evaluation \& Demonstration}
\label{sec:eval}
We evaluate the add-on components for system completeness, such as 2D and 3D vision, hardware configurations, and their performances. Other components such as mapping, localization, navigation, and planning are well-supported ROS packages: 2D Navigation Stack and MoveIt!.

\subsection{Data Preparation}
To verify the correctness of our proposed method, we first collect data from three in-lab objects: cones, cubes, and spheres. We then label them with annotations in Pascal VOC format and split our custom dataset into three sets: training set (70\%), validation set (20\%), and test set (10\%).

\begin{figure}[t]
    \centering
    \includegraphics[width=1.00\linewidth]{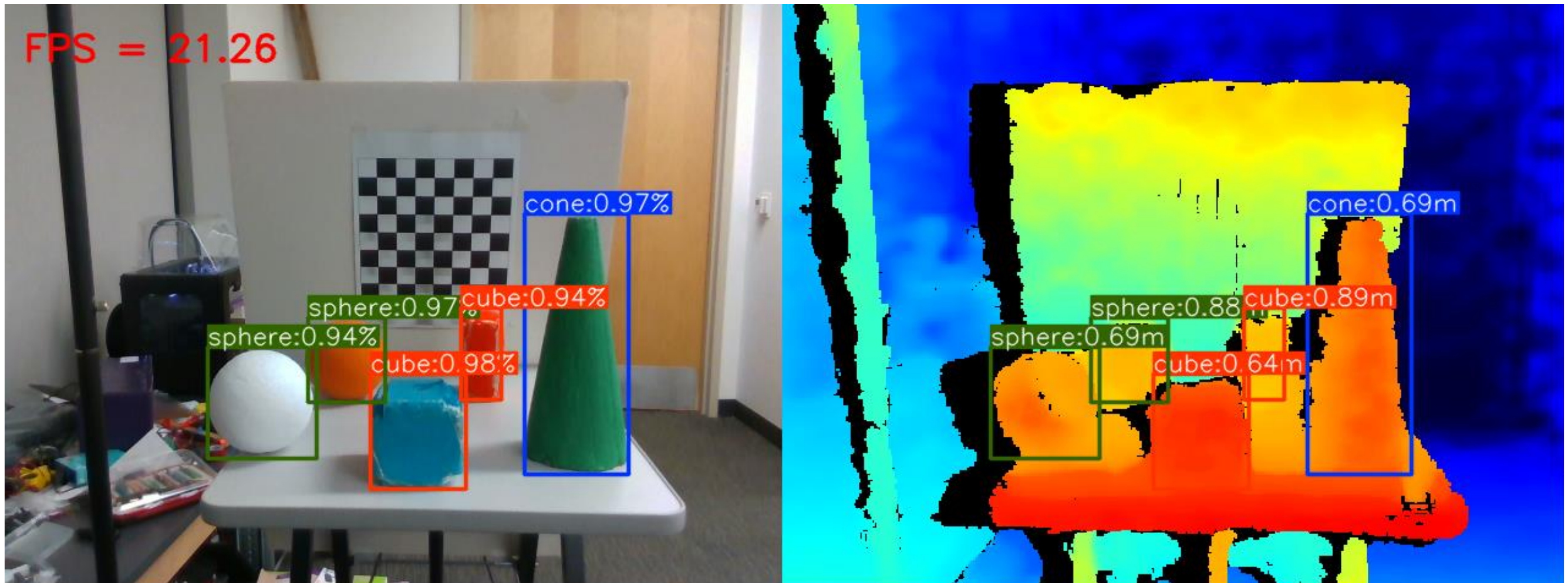}
    \caption{Experiment on simultaneous object detection using YOLOv7-tiny and depth estimation on Intel RS D435i camera.}
    \label{fig:depth_detection}
    \vspace{-15pt}
\end{figure}

\subsection{Evaluation Metrics}
To evaluate how well the transfer learning strategies perform during the training periods, we calculate the validation loss, average precision (AP), and mean average precision (mAP) for MobileNetv1 \citetk{howard2017mobilenets}, MobileNetv2 \citetk{sandler2018mobilenetv2}, SqueezeNet \citetk{iandola2016squeezenet}, VGG-16 \citetk{simonyan2014very}, and YOLOv7 \citetk{wang2023yolov7}. We train each model with three different strategies, as illustrated in Fig. \ref{fig:transfer_learning_strategies}. Data augmentation is used in a preprocessing procedure to enrich the training dataset, including rotation, cropping, and color distortion. We also evaluate the detection performance on test sets using AP and mAP calculated based on multiple intersections over union (IoU) thresholds.

For each test image, the IoU is defined as $IoU = {(B_p\cap B_t)}/{(B_p\cup B_t)}$ where $B_p$ and $B_t$ are predicted bounding box and ground truth bounding box, respectively.  If an IoU of a prediction is greater or equal to the predefined threshold value, that prediction is classified as a true positive (TP); otherwise, it is counted as a false positive (FP). Since the testing stage produces multiple detections among the classes, we first sort the confidence scores of all predictions in descending order as we need the trend of recall scores ($P$) to ascend while the trend of recall scores ($R$) descends in the P-R curve. Hence, we compute $P$ and $R$ for predictions of proposal performance for each class, $P = cTP/(cTP + cFP)$ and $R = cTP/GT$, where $cTP$ and $cFP$ denote cumulative TP and cumulative FP, respectively, and $GT$ is each class's ground truth. Thus, the AP for each class is computed as $AP_j = \int_{0}^{1} \text{p}(r)$, where $j$ denotes for class "\textit{j}". However, the values we obtain are discrete; we defined the interpolated area under the P-curve along the R-axis in the P-R curve in terms of recall levels, $r_j$, as:
\begin{equation}
    AP_j = \sum_{i=1}^{n-1} \left(r_{i+1} - r_i \right) \cdot \max_{r'_{i+1} \geq r_{i+1}} p(r'_{i+1})
\end{equation}

\begin{figure}[t]
    \centering
    \includegraphics[width=1.00\linewidth]{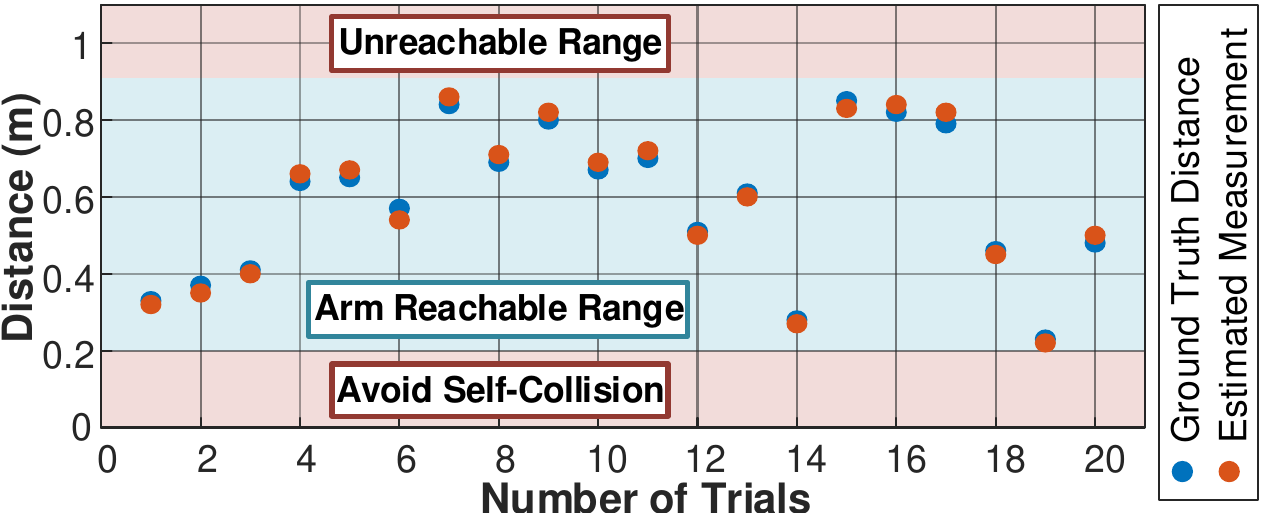}
    \caption{Estimated depth measurements compared to ground truth distances from kinematic transformation.}
    \label{fig:depth_estimation}
    \vspace{-15pt}
\end{figure}

\begin{figure*}[t]
    \includegraphics[width=1.00\textwidth]{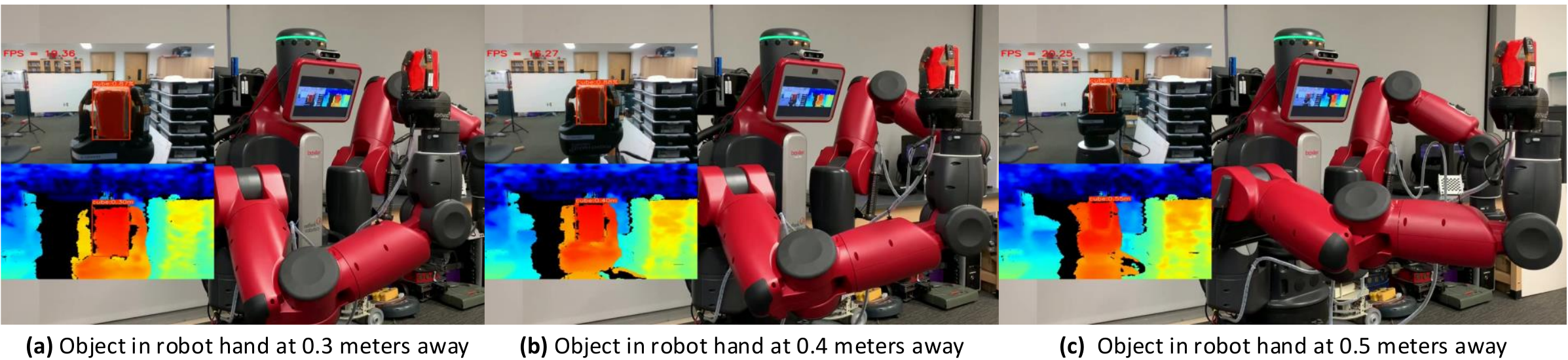}
    \centering
    \caption{Experiment setup of the Baxter robot simultaneously grasping the target-detected object (cube) and estimating its depth at \textbf{(a)} 0.3 meters, \textbf{(b)} 0.4 meters, and \textbf{(c)} 0.5 meters away from the mounted camera, respectively.}
    \vspace{-7pt}
    \label{fig:experiment_setup}
\end{figure*}

\begin{figure*}[t]
    \includegraphics[width=1.00\textwidth]{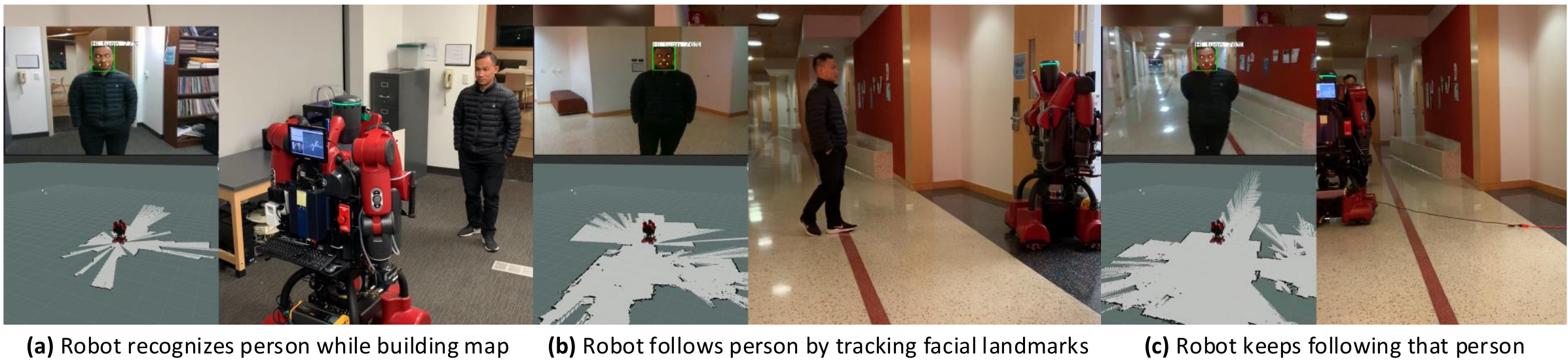}
    \centering
    \caption{Experiment setup of the Baxter robot \textbf{(a)} simultaneously recognizing the target person by tracking facial landmarks and building the environment map, \textbf{(b)} starting following the tracked person, and \textbf{(c)} keeping following that person.}
    \vspace{-15pt}
    \label{fig:experiment_slam}
\end{figure*}

The IoU thresholds range from 0.01 to 1.00 with a step of 0.01. After evaluating detection proposals on all IoU thresholds, we calculate the mAP for each model as below:
\begin{equation}
    mAP = \frac{\sum_{k=1}^{c}AP_k}{c}
\end{equation}
where $c$ is the number of classes in our training set. 

The AP and mAP results on multiple IoU thresholds ranging from 0.01 to 1.00 with a step of 0.01 are shown in Fig. \ref{fig:ap_and_map}.

\subsection{Results}
\indent \textbf{Training Performance}: We train MobileNetv1, MobileNetv2, SqueezeNet, VGG-16, and YOLOv7 on the NVIDIA GTX 3090 (24 GB) GPU with the three strategies mentioned in Sec. \ref{sec:perception} with 1000 epochs. The trained models start to converge at the $200^{th}$ epoch and finally converge at  the $800^{th}$ epoch, taking approximately 39 minutes. Note that YOLOv7 is trained with a different loss function than other models, so we only compare YOLOv7 in prediction performance. Fine-tuning gives the lowest loss among the three methods, and VGG-16 obtains the best results among models. Indeed, VGG-16 learns the best with fine-tuning transfer learning strategy, as shown in Fig. \ref{fig:valid_loss}.

\indent \textbf{Testing Performance}: We test detectors on commodity computers using Intel processors (i.e., Core i3-3217U and HD Graphics 4000). Like the training process, the fine-tuning strategy gives the highest accuracy, while the feature extraction transfer learning strategy gives the best result except for YOLOv7, as shown in Fig. \ref{fig:ap_and_map}. The feature extraction transfer learning strategy performs better than the randomly initialized weights strategy regarding YOLOv7. When detecting a sphere, there is a slightly different in the precision between feature extraction transfer and fine-tuning transfer strategies. Fig. \ref{fig:ap_and_map} also reveals that the source feature extraction in YOLOv7 works well with objects in our target domain, while other models fail to extract features from objects in our target domain. Lastly, YOLOv7 achieves the highest precision at the maximum IoU threshold, while feature extraction transfer learning does not work for SqueezeNet.

\begin{table}[h]
    \centering
    \setlength\tabcolsep{0.6pt}
    \begin{tabular}{|c |c |c |c |c |c |} 
        \hline \textbf{Network} & \textbf{\#Params} & \textbf{CPU} & \textbf{\small{Intel GPU}} & \textbf{VPU}\\
        \hline \footnotesize{MobileNetv1} & \footnotesize{6,883,296} & \footnotesize{14.87 $\pm$ 0.12}  & \footnotesize{19.37 $\pm$ 0.23} & \footnotesize{11.74 $\pm$ 0.07} \\
        \hline \footnotesize{MobileNetv2} & \footnotesize{3,087,328} &  \footnotesize{17.35 $\pm$ 0.19}  & \footnotesize{19.96 $\pm$ 0.22} & \footnotesize{10.15 $\pm$ 0.05} \\
        \hline \footnotesize{SqueezeNet} & \footnotesize{1,639,648} & \footnotesize{18.35 $\pm$ 0.22}  & \footnotesize{22.53 $\pm$ 0.28} & \footnotesize{14.82 $\pm$ 0.11} \\
        \hline \footnotesize{VGG-16} & \footnotesize{24,013,744} & \footnotesize{2.49 $\pm$ 0.01}  & \footnotesize{5.15 $\pm$ 0.02} & \footnotesize{2.22 $\pm$ 0.005} \\
        \hline \footnotesize{YOLOv7-tiny} & \footnotesize{6,652,669}  & \footnotesize{12.59 $\pm$ 0.07} & \footnotesize{21.47 $\pm$ 0.22} & \footnotesize{13.67 $\pm$ 0.08} \\
        \hline
    \end{tabular}
    \caption{Detection performance of models in frames per second (fps) on different hardware configurations (implemented using OpenVINO) with a confidence level of 95\%.}
    \label{table:hardware_configuration}
\end{table}

\indent \textbf{Hardware Configuration}: We run each detection model on CPU, GPU (Intel), and VPU (Intel NCS2) for $n = 300$ samples and calculate confidence intervals: $CI = \overline{fps} \pm z_{\alpha/2}\cdot ({\sigma}/{\sqrt{n}})$, where $\overline{fps}$ is mean frame rate (fps), $\sigma$ is the standard deviation, and $z$ is the confidence level value of $\alpha = 95\%$. The test is implemented using OpenVINO, which enables ML models to run on Intel onboard GPU. We also test on a VPU interfacing with a computer via USB. Onboard GPU outperforms other computing devices in terms of frame rate. MoblileNetv2 outperforms YOLOv7 when being tested on the CPU but underperforms YOLOv7 on the onboard GPU and VPU. Lastly, the VPU maintains the most stable performance due to its lowest variance (Table \ref{table:hardware_configuration}). The reason for obtaining the low variance on VPU is it does not share the workload like CPU or onboard GPU.

\indent \textbf{Depth Estimation}: We use techniques described in Eq. \ref{eq:depth_estimation} to estimate depth of detected objects (Fig. \ref{fig:depth_detection}). To generate ground truth distances, we teach the robot to grasp and hold an object in its gripper and then calculate the distance between the robot and that object using the kinematic transformation. Fig. \ref{fig:depth_estimation} shows that the minimum error is 1.00 cm, the maximum error is 3.00 cm, and the mean error is 1.75 cm.


\subsection{Demonstration}
The demonstration video includes two scenarios: (1) the robot grasps an object while estimating the depth of the detected object (Fig. \ref{fig:experiment_setup}), and (2) the robot performs SLAM while following a person using a face recognition module (Fig. \ref{fig:experiment_slam}) running on the Intel NC2: \url{https://youtu.be/q4oz9Rixbzs}.

%% file: 07_limitations.tex
\section{Current Limitations}
As our implementation is tight to ROS APIs and OpenVINO, the proposed design is only compatible with this software framework. Scaling to other software systems needs to rewrite the adaptation layer but can retain the other components because these are independent of this framework.  

%% file: 08_conclusions.tex
\section{Conclusions \& Future Works}
This work proposes a software-hardware framework for mobile cobots focusing on building and optimizing 2D and 3D perception with multiple commodity hardware. We build the framework on top of multiple ROS distributions, Linux versions, and OpenVINO. For design purposes, the framework can support multiple hardware and find the optimal configurations for input devices/sensors and computing devices. We mathematically model our transfer learning strategies and evaluate them on different computing devices. We then tested our framework on a 7-DOF two-arm Baxter robot with 2D detection and 3D depth estimation. An end-user application is also introduced for system completeness to facilitate software reusability. We reserve advanced techniques in robot 3D perception, such as segmentation, detection, and recognition, from a point cloud perspective for future works.

%% file: 09_acknowledgement.tex
\section{Acknowlegment}
We would like to thank Christopher Collander (LEARN Lab) for his initial support in this project.

%% file: 00_main.bbl
\begin{thebibliography}{10}
\providecommand{\url}[1]{#1}
\csname url@samestyle\endcsname
\providecommand{\newblock}{\relax}
\providecommand{\bibinfo}[2]{#2}
\providecommand{\BIBentrySTDinterwordspacing}{\spaceskip=0pt\relax}
\providecommand{\BIBentryALTinterwordstretchfactor}{4}
\providecommand{\BIBentryALTinterwordspacing}{\spaceskip=\fontdimen2\font plus
\BIBentryALTinterwordstretchfactor\fontdimen3\font minus
  \fontdimen4\font\relax}
\providecommand{\BIBforeignlanguage}[2]{{%
\expandafter\ifx\csname l@#1\endcsname\relax
\typeout{** WARNING: IEEEtran.bst: No hyphenation pattern has been}%
\typeout{** loaded for the language `#1'. Using the pattern for}%
\typeout{** the default language instead.}%
\else
\language=\csname l@#1\endcsname
\fi
#2}}
\providecommand{\BIBdecl}{\relax}
\BIBdecl

\bibitem{hsiao2009reactive}
K.~Hsiao, P.~Nangeroni, M.~Huber, A.~Saxena, and A.~Y. Ng, ``Reactive grasping
  using optical proximity sensors,'' in \emph{2009 IEEE International
  Conference on Robotics and Automation}.\hskip 1em plus 0.5em minus
  0.4em\relax IEEE, 2009, pp. 2098--2105.

\bibitem{hmedan2022adapting}
B.~Hmedan, D.~Kilgus, H.~Fiorino, A.~Landry, and D.~Pellier, ``Adapting cobot
  behavior to human task ordering variability for assembly tasks,'' in
  \emph{The International FLAIRS Conference Proceedings}, vol.~35, 2022.

\bibitem{vice2022leveraging}
J.~Vice, G.~Sukthankar, and P.~K. Douglas, ``Leveraging evolutionary algorithms
  for feasible hexapod locomotion across uneven terrain,'' \emph{arXiv preprint
  arXiv:2203.15948}, 2022.

\bibitem{macenski2022robot}
S.~Macenski, T.~Foote, B.~Gerkey, C.~Lalancette, and W.~Woodall, ``Robot
  operating system 2: Design, architecture, and uses in the wild,''
  \emph{Science Robotics}, vol.~7, no.~66, p. eabm6074, 2022.

\bibitem{biondi2019safe}
A.~Biondi, F.~Nesti, G.~Cicero, D.~Casini, and G.~Buttazzo, ``A safe, secure,
  and predictable software architecture for deep learning in safety-critical
  systems,'' \emph{IEEE Embedded Systems Letters}, vol.~12, no.~3, pp. 78--82,
  2019.

\bibitem{nazarova2021cobotar}
E.~Nazarova, O.~Sautenkov, M.~A. Cabrera, J.~Tirado, V.~Serpiva,
  V.~Rakhmatulin, and D.~Tsetserukou, ``Cobotar: interaction with robots using
  omnidirectionally projected image and dnn-based gesture recognition,'' in
  \emph{2021 IEEE International Conference on Systems, Man, and Cybernetics
  (SMC)}.\hskip 1em plus 0.5em minus 0.4em\relax IEEE, 2021, pp. 2590--2595.

\bibitem{dang2023perfc}
T.~Dang, K.~Nguyen, and M.~Huber, ``Perfc: An efficient 2d and 3d perception
  software-hardware framework for mobile cobot,'' in \emph{The International
  FLAIRS Conference Proceedings}, vol.~36, 2023.

\bibitem{rendiniello2020flexible}
A.~Rendiniello, A.~Remus, I.~Sorrentino, P.~K. Murali, D.~Pucci, M.~Maggiali,
  L.~Natale, S.~Traversaro, E.~Villagrossi, A.~Polo \emph{et~al.}, ``A flexible
  software architecture for robotic industrial applications,'' in \emph{2020
  25th IEEE International Conference on Emerging Technologies and Factory
  Automation (ETFA)}, vol.~1.\hskip 1em plus 0.5em minus 0.4em\relax IEEE,
  2020, pp. 1273--1276.

\bibitem{qureshi2019motion}
A.~H. Qureshi, A.~Simeonov, M.~J. Bency, and M.~C. Yip, ``Motion planning
  networks,'' in \emph{2019 International Conference on Robotics and Automation
  (ICRA)}.\hskip 1em plus 0.5em minus 0.4em\relax IEEE, 2019, pp. 2118--2124.

\bibitem{pinto2016supersizing}
L.~Pinto and A.~Gupta, ``Supersizing self-supervision: Learning to grasp from
  50k tries and 700 robot hours,'' in \emph{2016 IEEE international conference
  on robotics and automation (ICRA)}.\hskip 1em plus 0.5em minus 0.4em\relax
  IEEE, 2016, pp. 3406--3413.

\bibitem{veloso2015cobots}
M.~Veloso, J.~Biswas, B.~Coltin, and S.~Rosenthal, ``Cobots: Robust symbiotic
  autonomous mobile service robots,'' in \emph{Twenty-fourth international
  joint conference on artificial intelligence}, 2015.

\bibitem{bellotto2008multisensor}
N.~Bellotto and H.~Hu, ``Multisensor-based human detection and tracking for
  mobile service robots,'' \emph{IEEE Transactions on Systems, Man, and
  Cybernetics, Part B (Cybernetics)}, vol.~39, no.~1, pp. 167--181, 2008.

\bibitem{kingston2019exploring}
Z.~Kingston, M.~Moll, and L.~E. Kavraki, ``Exploring implicit spaces for
  constrained sampling-based planning,'' \emph{The International Journal of
  Robotics Research}, vol.~38, no. 10-11, pp. 1151--1178, 2019.

\bibitem{ichnowski2020gomp}
J.~Ichnowski, M.~Danielczuk, J.~Xu, V.~Satish, and K.~Goldberg, ``Gomp:
  Grasp-optimized motion planning for bin picking,'' in \emph{2020 IEEE
  International Conference on Robotics and Automation (ICRA)}.\hskip 1em plus
  0.5em minus 0.4em\relax IEEE, 2020, pp. 5270--5277.

\bibitem{chitta2012moveit}
S.~Chitta, I.~Sucan, and S.~Cousins, ``Moveit![ros topics],'' \emph{IEEE
  Robotics \& Automation Magazine}, vol.~19, no.~1, pp. 18--19, 2012.

\bibitem{liu2016ssd}
W.~Liu, D.~Anguelov, D.~Erhan, C.~Szegedy, S.~Reed, C.-Y. Fu, and A.~C. Berg,
  ``Ssd: Single shot multibox detector,'' in \emph{Computer Vision--ECCV 2016:
  14th European Conference, Amsterdam, The Netherlands, October 11--14, 2016,
  Proceedings, Part I 14}.\hskip 1em plus 0.5em minus 0.4em\relax Springer,
  2016, pp. 21--37.

\bibitem{wang2023yolov7}
C.-Y. Wang, A.~Bochkovskiy, and H.-Y.~M. Liao, ``Yolov7: Trainable
  bag-of-freebies sets new state-of-the-art for real-time object detectors,''
  in \emph{Proceedings of the IEEE/CVF Conference on Computer Vision and
  Pattern Recognition}, 2023, pp. 7464--7475.

\bibitem{mao20223d}
J.~Mao, S.~Shi, X.~Wang, and H.~Li, ``3d object detection for autonomous
  driving: a review and new outlooks,'' \emph{arXiv preprint arXiv:2206.09474},
  2022.

\bibitem{qi2017pointnet}
C.~R. Qi, H.~Su, K.~Mo, and L.~J. Guibas, ``Pointnet: Deep learning on point
  sets for 3d classification and segmentation,'' in \emph{Proceedings of the
  IEEE conference on computer vision and pattern recognition}, 2017, pp.
  652--660.

\bibitem{dang2022iotree}
T.~Dang, T.~Tran, K.~Nguyen, T.~Pham, N.~Pham, T.~Vu, and P.~Nguyen, ``iotree:
  a battery-free wearable system with biocompatible sensors for continuous tree
  health monitoring,'' in \emph{Proceedings of the 28th Annual International
  Conference on Mobile Computing And Networking}, 2022, pp. 769--771.

\bibitem{grisetti2005improving}
G.~Grisetti, C.~Stachniss, and W.~Burgard, ``Improving grid-based slam with
  rao-blackwellized particle filters by adaptive proposals and selective
  resampling,'' in \emph{Proceedings of the 2005 IEEE international conference
  on robotics and automation}.\hskip 1em plus 0.5em minus 0.4em\relax IEEE,
  2005, pp. 2432--2437.

\bibitem{zhuang2020comprehensive}
F.~Zhuang, Z.~Qi, K.~Duan, D.~Xi, Y.~Zhu, H.~Zhu, H.~Xiong, and Q.~He, ``A
  comprehensive survey on transfer learning,'' \emph{Proceedings of the IEEE},
  vol. 109, no.~1, pp. 43--76, 2020.

\bibitem{howard2017mobilenets}
A.~G. Howard, M.~Zhu, B.~Chen, D.~Kalenichenko, W.~Wang, T.~Weyand,
  M.~Andreetto, and H.~Adam, ``Mobilenets: Efficient convolutional neural
  networks for mobile vision applications,'' \emph{arXiv preprint
  arXiv:1704.04861}, 2017.

\bibitem{sandler2018mobilenetv2}
M.~Sandler, A.~Howard, M.~Zhu, A.~Zhmoginov, and L.-C. Chen, ``Mobilenetv2:
  Inverted residuals and linear bottlenecks,'' in \emph{Proceedings of the IEEE
  conference on computer vision and pattern recognition}, 2018, pp. 4510--4520.

\bibitem{iandola2016squeezenet}
F.~N. Iandola, S.~Han, M.~W. Moskewicz, K.~Ashraf, W.~J. Dally, and K.~Keutzer,
  ``Squeezenet: Alexnet-level accuracy with 50x fewer parameters and< 0.5 mb
  model size,'' \emph{arXiv preprint arXiv:1602.07360}, 2016.

\bibitem{simonyan2014very}
K.~Simonyan and A.~Zisserman, ``Very deep convolutional networks for
  large-scale image recognition,'' \emph{arXiv preprint arXiv:1409.1556}, 2014.

\bibitem{kim_gateway}
J.~Kim, T.~Dang, J.~Jeon, and B.~Yeom, ``Design of a seamless gateway for Mechatrolink?,'' \emph{2013 IEEE International Conference on Industrial Technology (ICIT)}, 2013.


\end{thebibliography}
